\documentclass[letterpaper, 10 pt, conference]{ieeeconf}  

\IEEEoverridecommandlockouts                              

\usepackage{cite}
\usepackage{amsmath,amssymb,amsfonts}
\usepackage{algorithmic}
\usepackage{graphicx}
\usepackage{textcomp}
\usepackage{xcolor}
\usepackage{booktabs}
\usepackage{multirow}
\usepackage{hyperref}
\usepackage{times}
\title{\LARGE \bf
Impact of Static Friction on Sim2Real in Robotic Reinforcement Learning
}

\author{ Xiaoyi Hu$^{1}$, Qiao Sun$^{1}$, Bailin He$^{1}$, Haojie Liu$^{1}$, Xueyi Zhang$^{1}$, Chunpeng lu$^{1}$ and Jiangwei Zhong$^{1}$
\thanks{Corresponding author: Jiangwei Zhong}
\thanks{$^{1}$X. Hu, Q. Sun, B. He, H. Liu, X. Zhang, C. Lu and J.Zhong  are with the Smart Devices \& Solutions Lab, Lenovo Research Shanghai, Shanghai 200003, China 
{\tt\small zhongjw@lenovo.com}}%
}

\begin{document}

\maketitle
\thispagestyle{empty}
\pagestyle{empty}

\begin{abstract}
In robotic reinforcement learning, the Sim2Real gap remains a critical challenge.
However, the impact of Static friction on Sim2Real has been underexplored.
Conventional domain randomization methods typically exclude Static friction from their parameter space.
In our robotic reinforcement learning task, such conventional domain randomization approaches resulted in significantly underperforming real-world models.
To address this Sim2Real challenge, we employed Actuator Net as an alternative to conventional domain randomization.
While this method enabled successful transfer to flat-ground locomotion, it failed on complex terrains like stairs.
To further investigate physical parameters affecting Sim2Real in robotic joints, we developed a control-theoretic joint model and performed systematic parameter identification.
Our analysis revealed unexpectedly high friction-torque ratios in our robotic joints.
To mitigate its impact, we implemented Static friction-aware domain randomization for Sim2Real.
Recognizing the increased training difficulty introduced by friction modeling, we proposed a simple and novel solution to reduce learning complexity.
To validate this approach, we conducted comprehensive Sim2Sim and Sim2Real experiments comparing three methods: conventional domain randomization (without Static friction), Actuator Net, and our Static friction-aware domain randomization.
All experiments utilized the Rapid Motor Adaptation (RMA) algorithm.
Results demonstrated that our method achieved superior adaptive capabilities and overall performance.

\end{abstract}

\section{INTRODUCTION}
In the realm of robotic reinforcement learning, Sim2Real problem is a consistently significant challenge.
This gap often results in robots that perform admirably in simulated environments but underperform in real-world settings.
While it is possible for robots to learn directly from real-world sampling, this approach is fraught with drawbacks, including potential harm to the robot and exorbitant costs \cite{gu2016deep}.
To enhance the data efficiency of reinforcement learning, Model-Based Reinforcement Learning (MBRL) presents a viable alternative \cite{moerland2023model}.
However, MBRL shares a common issue with Model Predictive Control (MPC), where the efficacy of learning is heavily contingent on the accuracy of the environment model and struggles to accommodate unpredictable external disturbances during robot operation \cite{huang2020model}, \cite{ernst2008reinforcement}.
It is due to these challenges that Model-free Reinforcement Learning (MFRL) has become the predominant learning method in robotic reinforcement learning \cite{singh2022reinforcement}.
Nonetheless, MFRL is inherently data-inefficient, prompting the exploration of simulation environments as a means to emulate real-world robotics.
In such simulations, robots can bypass direct interaction with the physical environment and significantly boost data collection efficiency through parallelization strategies \cite{makoviychuk2021isaac}, \cite{akkaya2019solving}.
Yet, in practical applications, disparities between simulated and real environments often arise due to computational complexity and technical limitations.
These discrepancies, though not obvious, frequently lead to suboptimal robot performance upon deployment \cite{dulac2021challenges}.

To address this challenge, a series of methods have been proposed to bridge the Sim2Real gap, many of which have demonstrated promising results.
Among these, domain randomization stands out as a widely used and effective technique\cite{tobin2017domain}.
By randomizing the physical properties of the robot, domain randomization significantly enhances the model's generalization capabilities.
However, excessive randomization can substantially increase training difficulty, potentially degrading model performance.
Therefore, it is crucial to analyze which factors significantly impact Sim2Real and to what extent.
To tackle this issue, we systematically controlled for several factors that may influence Sim2Real and conducted deployment tests on our Saturn Lite robot.
During deployment, we observed that, unlike most reinforcement learning deployment studies, randomizing only the rotor inertia, viscous friction, Kp, and Kd of joint motors resulted in poor performance.
Only by incorporating Static friction into the randomization process could we achieve satisfactory robot performance.
To quantitatively evaluate the impact of Static friction, we employed three approaches: Domain Randomization without Static friction, Actuator Net, and Domain Randomization with Static friction.
Additionally, based on control-theoretic analysis, we identified key parameters for randomization in Domain Randomization and analyzed the impact of Static friction in robot performance.

In our robotic reinforcement learning training tasks, the effectiveness of Sim2Real was primarily assessed by evaluating the Saturn Lite's walking and terrain adaptation capabilities.
The Saturn Lite robot, a hexapod developed by Lenovo's Shanghai Research Institute, boasts greater stability compared to mainstream quadruped robots due to its tripod gait.
However, the additional legs also introduce more degrees of freedom, complicating the robot's gait.
Constraining the robot's gait at its footholds is a feasible strategy, but this approach limits the robot to walking on flat surfaces and eliminates its ability to self-adjust its footholds.

\begin{figure}\centering
    \vspace{0.3cm}
    \includegraphics[scale=0.62]{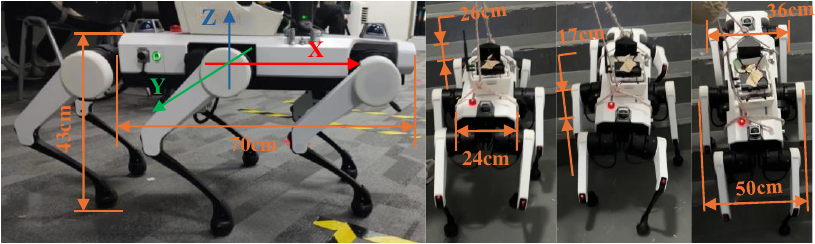}
    \caption{
        The Saturn Lite robot utilizing a reinforcement learning algorithm for locomotion. The figure captures the robot's adaptive process across stairs. During terrain adaptation, the robot relies solely on proprioceptive sensor (such as IMU and joint encoder) without exteroceptive sensor (e.g., LiDAR or cameras).
    }
    \label{fig:saturn_lite}
\end{figure}

To enable the robot to autonomously adapt to various terrains using a tripod gait, we have proposed a novel reward function designed to incentivize such behavior.
The effectiveness of our method was validated through tests of the robot's ability to traverse stairs and flat terrain in the real world.
The configuration of our robot is shown in Fig. \ref{fig:saturn_lite}.

In summary, our contributions are as follows:
\begin{itemize}
    \item To the best of our knowledge, we are the first to systematically identify and demonstrate the impact of Static friction on robotic reinforcement learning. Through parameter identification and comparative Sim2Sim and Sim2Real experiments, we validated this hypothesis.

    \item By developing a second-order control model for the motors, we analyzed which domain randomization parameters can be excluded from randomization. Furthermore, we proposed a solution to address the effects of Static friction in reinforcement learning for robots with significant friction.

    \item Our methods were implemented on a novel hexapod robot, SaturnLite, showing its effectiveness to solve the static friction problem and resulting in better locomotion performance, compared to traditional methods.
\end{itemize}
\section{RELATED WORK}
The performance of Sim2Real is jointly influenced by simulation environments and reinforcement learning algorithms. To systematically analyze the Sim2Real challenge, this section critically examines the strengths and limitations of existing simulation platforms and reinforcement learning algorithms for legged robot training.
\subsection{Analysis of Existing Simulation platforms}\label{Analysis of Existing Simulation Environments}
Simulation platforms play a pivotal role in robotics research, yet significant variations exist in their simulation speed and accuracy.
Notably, the disparity in simulation speeds across different platforms can span three orders of magnitude (1,000-8,000x) \cite{Genesis}.
While such speed differences substantially impact training time costs, accelerated simulation often comes at the expense of reduced physical accuracy \cite{gu2024humanoid, erez2015simulation}.
This fundamental trade-off between efficiency and fidelity necessitates careful consideration when selecting simulation environments for reinforcement learning applications.

Current simulation architectures can be categorized by their parallelization strategies: CPU-based and GPU-based implementations.
Conventional CPU-based platforms (e.g., Webots \cite{michel2004cyberbotics}, PyBullet \cite{coumans2016pybullet}, RaiSim \cite{hwangbo2018per}, MuJoCo \cite{erez2015simulation}) require substantial computational resources to achieve parity with GPU-accelerated environments like Isaac Gym \cite{makoviychuk2021isaac} and Isaac Sim \cite{isaacsim}.
For instance, OpenAI's robotic cube-solving demonstration necessitated 64 NVIDIA V100 GPUs combined with 29,440 CPU cores for real-world deployment \cite{akkaya2019solving}, whereas equivalent computational throughput can be achieved using a single GPU in Isaac-based environments\cite{makoviychuk2021isaac}.
This dramatic reduction in training costs has propelled Isaac Gym and Isaac Sim to become the fastest-growing simulation platforms in reinforcement learning research \cite{kaup2024review}.

However, reduced computational expense does not inherently improve simulation accuracy.
The PhysX physics engine underlying Isaac platforms demonstrates inferior physical fidelity compared to MuJoCo's solver, potentially exacerbating the Sim2Real gap.
To quantify this discrepancy, researchers often employ a two-stage validation approach: models trained in Isaac environments are deployed in MuJoCo for evaluation\cite{gu2024humanoid}.
While this methodology eliminates physical safety risks by avoiding real-world interactions, it fails to fully address the fundamental Sim2Real gap.
This persistent challenge has motivated the development of various algorithms specifically designed to bridge this gap.

\subsection{Reinforcement Learning for Legged Robots and Sim2Real Solutions}\label{Reinforcement Learning for Legged Robots and Sim2Real Solutions}
Given that our training methodology relies on proprioceptive sensing (i.e., terrain recognition through foot-ground contact) and considering the limited research on hexapod robot reinforcement learning, this subsection focuses on proprioception-based reinforcement learning algorithms for quadruped robots and Sim2Real gap mitigation.

Compared to wheeled robots, legged robots necessitate simultaneous optimization for multi-terrain traversal capabilities, making terrain adaptability a critical requirement in its training.
However, conventional Domain Randomization (DR) methods—which typically randomize physical properties, sensor noise, and actuator dynamics to address Sim2Real discrepancies—face exacerbated learning challenges when combined with terrain adaptation demands, often resulting in suboptimal real-world performance.
To enhance real-world transferability, researchers have proposed innovative solutions.

Chen et al. introduced the concept of privileged observation to address complex training scenarios, where a student policy learns to imitate a teacher policy with access to privileged observations \cite{chen2020learning}.
This approach was successfully adapted to legged robots by Lee et al., who utilized terrain elevation maps and foot collision states as privileged observations to enable quadrupedal locomotion on complex terrains \cite{lee2020learning}.
Kumar et al. extended this work by developing an adaptation module for real-time terrain generalization \cite{kumar2021rma}.

To mitigate the computational overhead of two-stage training frameworks, Nahrendra et al. proposed Dream WaQ, which employs a Variational Autoencoder (VAE) to extract latent representations \cite{nahrendra2023dreamwaq}.
By concatenating these representations with non-privileged observations, the method achieves terrain adaptation within a single training phase.
Long et al. further enhanced robustness by integrating H$\infty$ control into reinforcement learning, formulating an adversarial framework between the policy and disturbance models to improve disturbance rejection capabilities \cite{long2024learning}.

While these methods predominantly leverage Domain Randomization, alternative approaches have demonstrated comparable efficacy in bridging the Sim2Real gap.
Notably, Hwangbo et al. developed Actuator Net, which collects real-world joint actuator data to simulate actuator dynamics in simulation \cite{hwangbo2019learning}.
By training policies with simulated actuator models, this method effectively addresses real-world actuator discrepancies without requiring hardware-in-the-loop testing.

\section{ROBOT CONFIGURATION AND SIMULATION TRAINING METHODOLOGY}
This section details the robotic platform used for deployment and the reinforcement learning (RL) training method.
\begin{figure}[t]
    \centering
    \vspace{0.25cm}
    \includegraphics[scale=0.28]{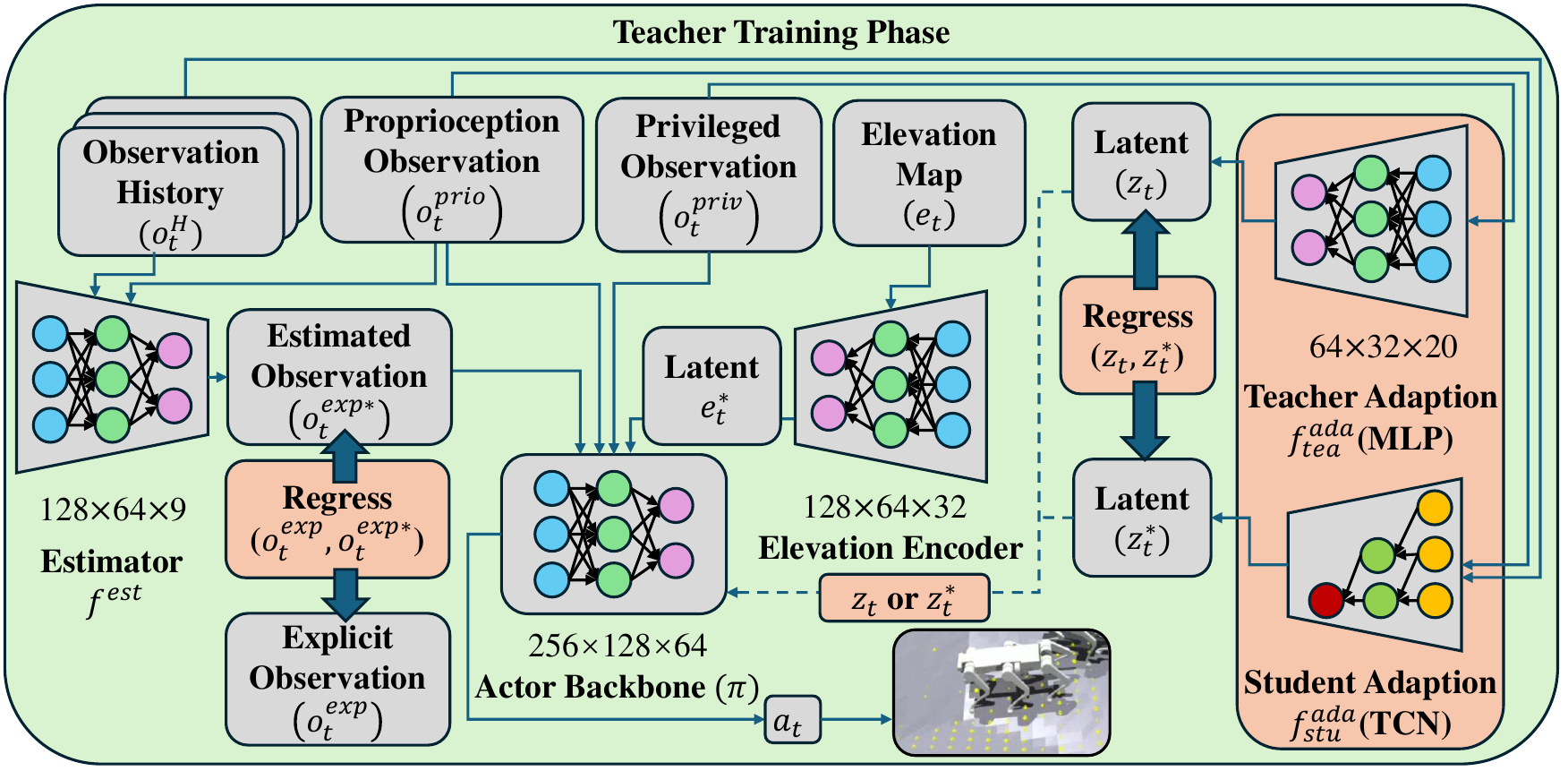}
    \caption{
        The training process for the teacher network.
        For simplicity, the critic network $V(s_t)$ is not shown in the figure, but its input $s_t$ includes all observations: $s_t = \{ o_t^\text{prio}, o_t^{H}, o_t^\text{priv}, o_t^\text{exp}, e_t \}$}.
    \label{fig:teacher_training}
\end{figure}
\subsection{Robot Configuration}
Our robotic platform is a hexapod robot, with each leg comprising three degrees of freedom (DoF) and a total body mass of 32 kg.
Compared to the widely used Go1 robot—a mainstream platform for RL deployment—our robot exhibits a significantly increased body mass while maintaining nearly identical joint torque capabilities.
The significant increase in the robot's mass is attributed to our enhancements in endurance.
This improvement were implemented to extend the operational duration, thereby expanding its practical applicability in real-world scenarios.
A comparative analysis of robot parameters between the Go1 and our Saturn Lite robot is presented in Table \ref{table:Robot Parameters Comparison}.
\begin{table}[]
    \caption{ROBOT PARAMETER COMPARISON}
    \label{table:Robot Parameters Comparison}
    \centering
    \begin{tabular}{@{}ccc@{}}
    \toprule
    \textbf{Parameter}       & \textbf{Go1} & \textbf{Saturn Lite (our)} \\ \midrule
    Total Body Mass (kg)     & 12    & 32          \\
    Hip Joint Torque (N·m)   & 23.7  & 25          \\
    Thigh Joint Torque (N·m) & 35.55 & 25          \\
    Shank Joint Torque (N·m) & 35.55 & 45          \\ 
    Endurance (Hours)        & 0.33  & 2           \\
    \bottomrule
    \end{tabular}
\end{table}
\subsection{Teacher Network Training}
Our network architecture is based on the Actor-Critic framework commonly used in reinforcement learning (RL), and we employ the Rapid Motor Adaptation (RMA) algorithm for training.
The RMA algorithm divides the training process into two phases: (1) the teacher phase, which has access to privileged observations, and (2) the student phase, which operates without privileged observations.
The teacher is trained using RL, while the student is trained through behavior cloning of the teacher.
Privileged observations refer to data that are unavailable in reality, such as terrain elevation maps around the robot or the robot's current mass.
These observations are introduced solely to enhance network performance during training.
Since such observations are inaccessible during real-world deployment, the student network learns to mimic the teacher's behavior to achieve comparable performance without relying on privileged observations.

During the training of the teacher network, the robot's state $s_t$ at time $t$ consists of five components: proprioceptive observations $o_t^\text{prio}$, observation history $o_t^{H}$, privileged observations $o_t^\text{priv}$, elevation maps $e_t$, and explicit observations $o_t^\text{exp}$.

The state $s_t$ and is formally defined as:  
\begin{equation}
s_t = \{ o_t^\text{prio}, o_t^{H}, o_t^\text{priv}, o_t^\text{exp}, e_t \}
\end{equation}
The proprioceptive observation $o_t^{prio}$ and the observation history $o_t^{H}$ are defined as follows:

\begin{equation}
\begin{gathered}
o_t^{prio} = \{ \omega_b^t, \phi_t^b, \psi_t^b, 0, com, 0, \theta_t, \dot{\theta}_t, a_{t-1}, 0 \}\\
o_t^{H} = \{ o_{t-H}^\text{prio}, o_{t+1-H}^\text{prio}, \dots, o_t^\text{prio} \}
\end{gathered}
\end{equation}
Here, $\omega_b^t$ is the angular velocity of the robot's base at time $t$;
$\phi_t^b$ and $\psi_t^b$ are the roll and pitch angles of the robot's base at time $t$, defined with respect to the robot's coordinate frame (see Figure \ref{fig:saturn_lite});
$0$ is a placeholder for fine-tuning when new observations are introduced;
$com$ is the command input determining the robot's movement direction;
$\theta_t$ and $\dot{\theta_t}$ are the joint angles and joint velocities at time $t$;
$a_{t-1}$ is the action executed at time $t-1$;
and ; and $H=10$ is the history length for $o^H_t$

The privileged observation $o_t^\text{priv}$ consists of parameters unavailable in real-world scenarios and subject to domain randomization, while the explicit observation $o_t^\text{exp}$ comprises quantities that require estimation:  
\begin{equation}
\begin{gathered}
o_t^\text{priv} = \{ m_b, f_g, I_j, B_j, f_j \}\\
o_t^\text{exp} = \{ v_t^b, c_t^\text{foot} \}
\end{gathered}
\end{equation}  
Here, $m_b$ denotes the mass of the robot's base;
$f_g$ represents the friction coefficient of the ground;
$I_j$, $B_j$, and $f_j$ correspond to the joint inertia, damping, and Static friction, respectively;
$v_t^b$ is the linear velocity of the robot's base;
and $c_t^\text{foot}$ indicates the contact state between the robot's feet and the ground.  
The elevation map $e_t$ provides terrain height data around the robot.  

During teacher training, since $o_t^\text{priv}$ and $o_t^\text{exp}$ are unavailable in real-world deployment, they are estimated using auxiliary networks.
The adaptation networks for the teacher $f_\text{tea}^\text{ada}$ and student $f_\text{stu}^\text{ada}$ are trained as follows:  
\begin{equation}
\begin{gathered}
z_t = f_\text{tea}^\text{ada}(o_t^\text{priv})\\
z_t^* = f_\text{stu}^\text{ada}(o_t^{H}, o_t^\text{prio})\\
L_\text{ada} = \text{MSE}(z_t, z_t^*)\\
\end{gathered}
\end{equation}  
Here, $z_t$ and $z_t^*$ are latent representations extracted by the teacher and student adaptation networks, respectively.  

For $o_t^\text{exp}$, an estimator network $f^\text{est}$ is used to predict its value based on $o_t^{H}$ and $o_t^\text{prio}$:  
\begin{equation}
\begin{gathered}
o_t^\text{exp*} = f^{est}(o_t^{H}, o_t^\text{prio})\\
L_\text{est} = \text{MSE}(o_t^\text{exp*}, o_t^\text{exp})
\end{gathered}
\end{equation}  

The actor network $\pi$ and critic network $V$ are defined as follows:  
\begin{equation}
\begin{gathered}
a_t = \pi(o_t^\text{exp*}, o_t^\text{prio}, z_t / z_t^*, e_t^*)\\
A_t = V(s_t)
\end{gathered}
\end{equation}  
Here, $e_t^*$ is the latent representation extracted from the elevation map $e_t$; $z_t$ and $z_t^*$ are latent representations from the teacher and student adaptation networks, which alternate during actor training; and $a_t$ is the position commands sent to the robot's joints.
The executed joint position $\theta_{pred}$ is computed relative to the robot's standing posture $\theta_{stand}$:  
\begin{equation}
\theta_\text{pred} = \theta_\text{stand} + a_t
\end{equation}  
Unlike the actor network, the critic network has access to all  observations $s_t$, including $o_t^\text{exp}$ and $o_t^\text{priv}$, which are unavailable in real-world scenarios.
$A_t$ represents the advantage function commonly used in RL training.  

The training process for the teacher network is illustrated in Figure \ref{fig:teacher_training}.

\subsection{Reward Fuction}

\begin{figure}
    \vspace{0.2cm}
    \centering
    \includegraphics[scale=0.38]{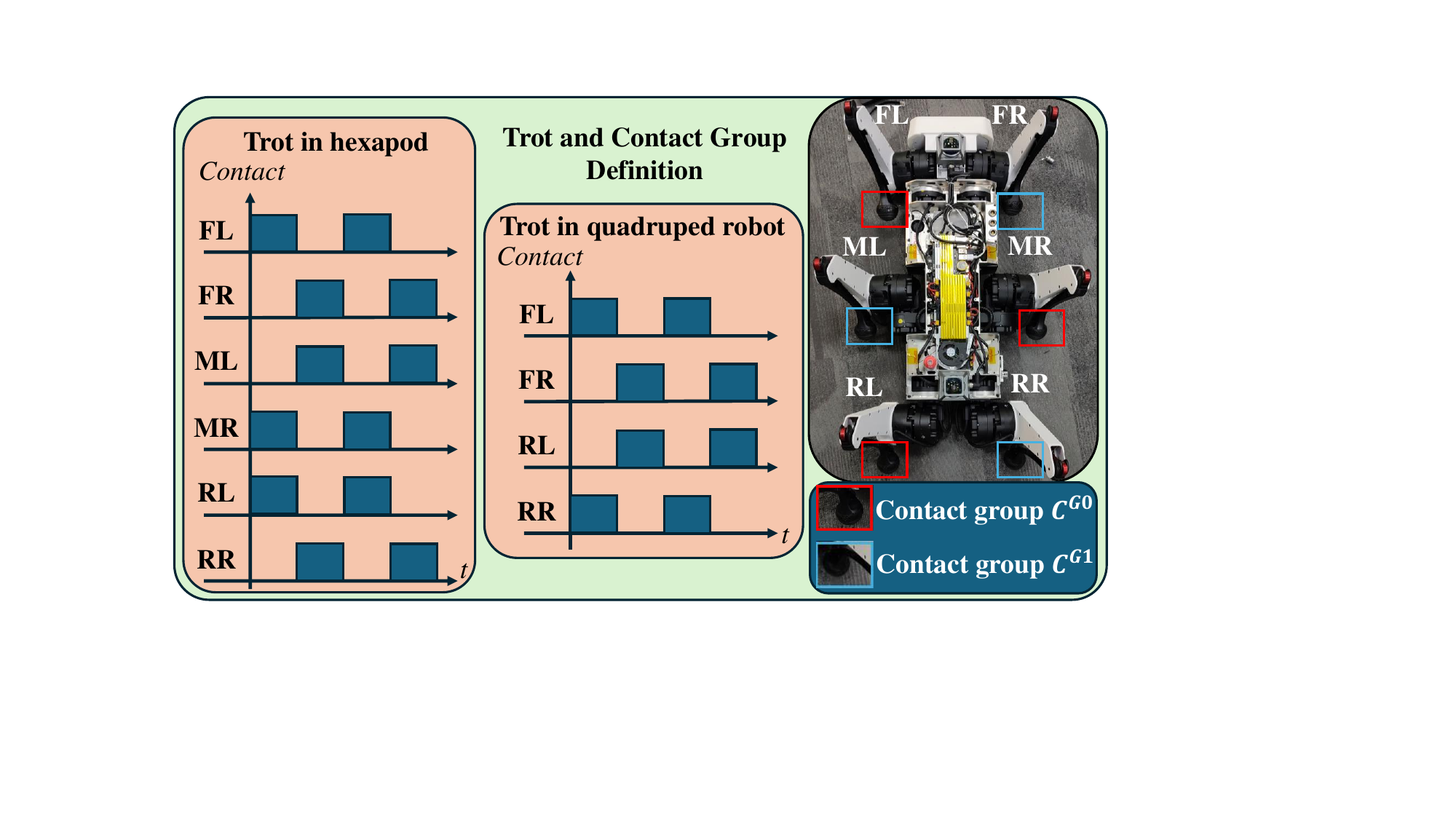}
    \caption{Trot gait for hexapod robots. For comparison, the gait cycle diagram for quadruped robots is also provided.}
    \label{fig:trot_definition}
\end{figure}

During locomotion, quadruped robots can exhibit various gaits, such as the \textbf{pronk gait} (all legs touching the ground or lifting off simultaneously) and the \textbf{trot gait} (legs on the same diagonal touching the ground while the other diagonal legs lift off).
For clarity, we define the pronk gait for hexapod robots similarly to quadruped robots, where all six legs touch the ground or lift off simultaneously.
However, the trot gait for hexapod robots is redefined, as illustrated in Figure \ref{fig:trot_definition}.  

The reward function for our hexapod robot is largely similar to those commonly used for quadruped robots.
However, during training and comparison, we observed that quadruped robots spontaneously adopt the trot gait under reinforcement learning guidance.
In contrast, applying the same reward function to our hexapod robot results in the pronk gait.
To constrain the robot to the desired trot gait, we propose the following reward function:  

\begin{equation}
\begin{aligned}
C^G_0 &= \{ FL, MR, RL \} \\
C^G_1 &= \{ FR, ML, RR \} \\
r_{\text{trot}} &= 
\begin{cases} 
1, & \text{if } (C_t^G = C^G_0) \oplus (C_t^G = C^G_1) \\
0, & \text{otherwise}
\end{cases} \\
r_{\text{unsync}} &= \left| \sum_{t=1}^H \left| C^G_0 \cap C_t^G \right| - \sum_{t=1}^H \left| C^G_1 \cap C_t^G \right| \right|
\end{aligned}
\end{equation}

Here, $FL, FR, ML, MR, RL, RR$ denote the front-left, front-right, middle-left, middle-right, rear-left, and rear-right legs, respectively.
To enforce the trot gait, we introduce the concept of \textbf{contact groups} $C^G$, where $C^G_0$ and $C^G_1$ represent the groups of legs that form the trot gait.
The contact group concept is further illustrated in Figure \ref{fig:trot_definition}. $r_\text{trot}$ ensures that only one contact group is active at any time, preventing the robot from adopting the pronk gait.  
$r_\text{unsync}$ ensures that the two contact groups are used equally over a period of time, preventing the robot from adopting a three-legged hopping gait.  
\subsection{Student Network}
After the teacher network is trained, a student network that does not rely on privileged observation $o_t^\text{priv} $ must be trained, as $o_t^\text{priv} $ is unavailable in real-world scenarios.
During the training of the student network, the estimator and student adaptation modules, which were already trained during the teacher phase, are loaded for fine-tuning.
Additionally, to ensure that the student's actions $a_t^* $ mimic the teacher's actions $a_t $, behavior cloning is performed between the student and teacher.
The corresponding loss function is defined as:
\begin{equation}
L_{\text{clone}} = \text{MSE}(a_t^*, a_t)
\end{equation}

\begin{figure}
\centering
\vspace{0.25cm}
\includegraphics[scale=0.42]{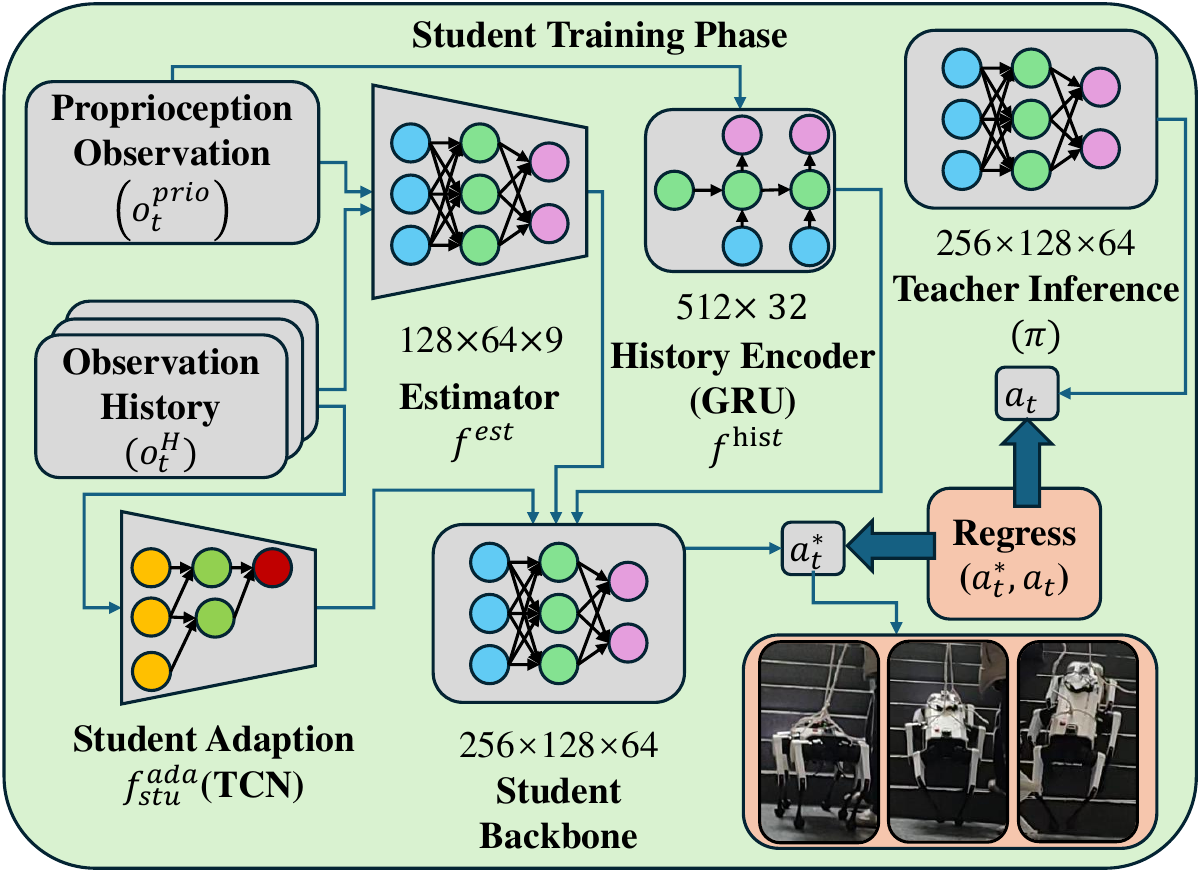}
\caption{Training process of the robot in the student phase.
During this phase, the student learns to mimic the actions of the teacher.}
\label{fig:student_training}
\end{figure}

Furthermore, since the student cannot access the surrounding elevation map, it cannot extract terrain feature directly.
To address this, we replace the terrain features $z_t^* $ extracted by the teacher's elevation encoder with features derived from the history encoder $f^{\text{hist}} $.
This allows the student to estimate terrain features based on the history of proprioceptive observations $o_t^{H} $.
The detailed training process for the student network is illustrated in Figure \ref{fig:student_training}.
After completing the student training phase, the resulting model can be deployed on physical hardware.

\begin{figure}[h]
    \centering
    \includegraphics[scale=0.38]{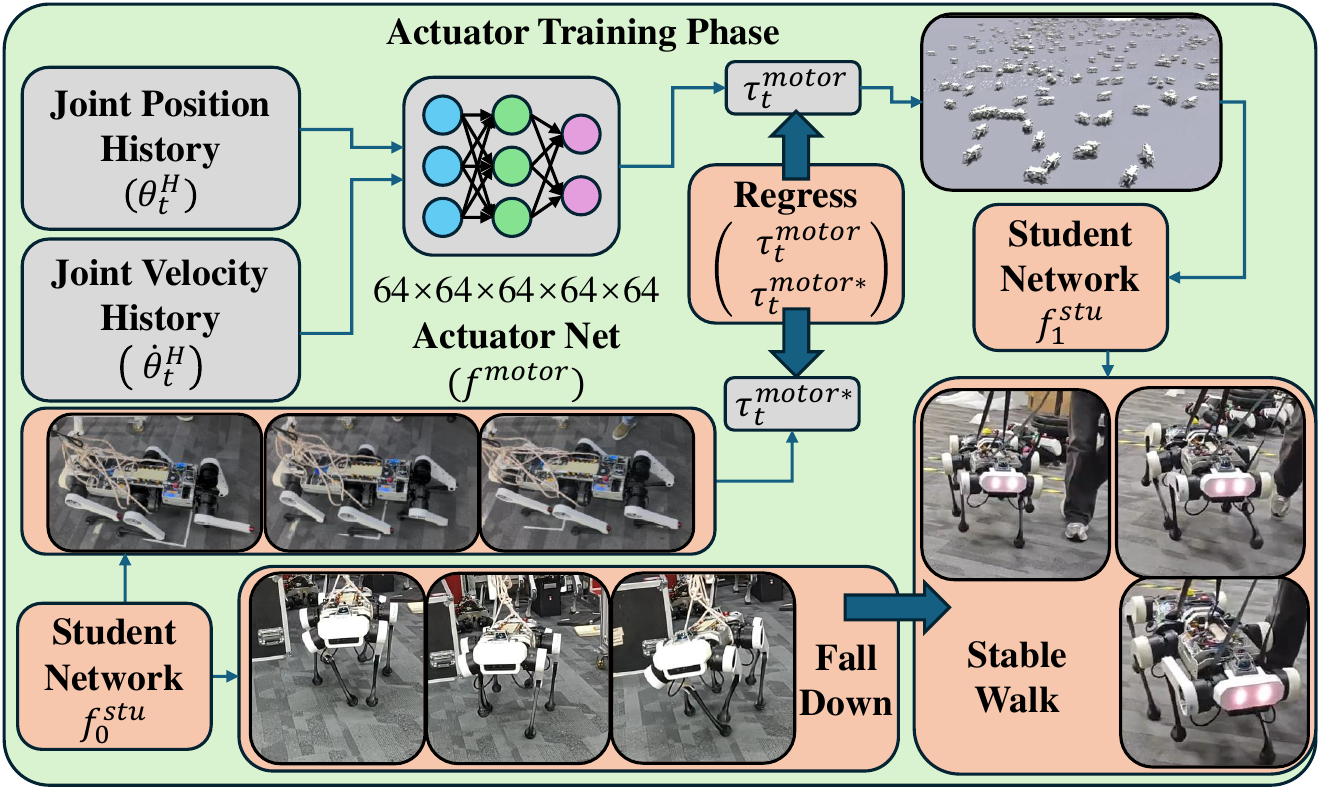}
    \caption{Sim2Real implementation based on Actuator Net.
    When using the student network $f_0^{\text{stu}}$ with Actuator Net, the robot frequently fell while walking forward, necessitating the use of a tether.
    After collecting walking data with $f_0^{\text{stu}}$, Actuator Net $f^{\text{motor}}$ was trained using this data.
    A new round of teacher-student training was then conducted using $f^{\text{motor}}$, resulting in a new student network $f_1^{\text{stu}}$.
    Walking based on $f_1^{\text{stu}}$ is stable but limited to flat ground.}
    \label{fig:actuator_traning}
\end{figure}
\section{SIM2REAL IMPLEMENTATION METHODS}
The aforementioned processes focus on network training in simulation environments.
However, robotic reinforcement learning also requires consideration of how well the model transfers to real-world robots.
During this process, we observed that commonly used domain randomization parameters performed poorly on our robot, often resulting in failure to walk properly.
To analyze and address this issue, we first employed Actuator Net $f^{\text{motor}}$ for model transfer.

\subsection{Sim2Real Based on Actuator Net}
The idea behind Actuator Net $f^{\text{motor}}$ is to collect data from real-world actuators, simulate the actuator model, and then use this model for training in the simulation environment.
To collect data for Actuator Net $f^{\text{motor}}$, we initially trained the robot using conventional domain randomization parameters and deployed the trained model on the physical robot.
Upon deployment, we found that the robot could only walk backward, while forward walking caused it to fall.
We used the backward walking data as the dataset to train Actuator Net $f^{\text{motor}}$.
The Actuator Net $f^{\text{motor}}$ can be expressed as:
\begin{equation}
\begin{gathered}
\tau_t^{\text{motor}} = f^{\text{motor}} \left( \theta_t^H, \dot{\theta}_t^H \right)\\
\theta_t^H = \{ \theta_t, \dots, \theta_{t-H} \}\\
\dot{\theta}_t^H = \{ \dot{\theta}_{t-1}, \dots, \dot{\theta}_{t-H} \}
\end{gathered}
\end{equation}

Here, $\theta_t^H$ and $\dot{\theta_t^H}$ represent the history of joint positions and joint velocities, respectively.
$\theta_t$ and $\dot{\theta_t}$ are the current joint position and velocity, while $\theta_{t-H} $ and $\dot{\theta}_{t-H}$ are the joint position and velocity $H$ steps ago.
In this case, $H = 3$.
The loss function $L_{\text{actuator}}$ for Actuator Net is defined as:
\begin{equation}
L_{\text{actuator}} = \text{MSE} \left( \tau_t^{\text{motor}}, \tau_t^{\text{motor}*} \right)
\end{equation}
Here, $\tau_t^{\text{motor}}$ is the joint torque predicted by Actuator Net, and $\tau_t^{\text{motor}*}$ is the actual joint torque measured during robot operation.
After training Actuator Net, we replaced the original PD controller in the simulation environment with Actuator Net and redeployed the model.
This allowed the robot to walk forward normally on flat ground.

The Sim2Real process based on Actuator Net is illustrated in Figure \ref{fig:actuator_traning}.
However, this approach only enables stable walking on flat ground and fails on stairs.
We hypothesize that this limitation arises because the Actuator Net training data only includes backward walking on flat ground.
Additionally, since the Actuator Net data was collected from a single robot, we speculate that the trained model may only be applicable to that specific robot.
Variations in joint across different robots could cause the deployment to fail.

\section{DOMAIN RANDOMIZATION IN ROBOT JOINTS BASED ON CONTROL THEORY ANALYSIS}
Since the Sim2Real implementation using Actuator Net only enables walking on flat ground and significantly increases training time due to iterative refinement, we analyzed the reasons why commonly used domain randomization parameters cause failures in our robot.
We systematically examined the physical parameters of the joints and proposed improvements based on this analysis.
To facilitate the analysis of joint parameters, we modeled the dynamics of a single joint $j$ during domain randomization as follows:
\begin{equation}
\begin{aligned}
I_j \ddot{\theta}_j(t) + B_j \dot{\theta}_j(t) = \tau_j(t) + f_j(t)\\
f_j(t) = 
\begin{cases} 
- b_j^c, & \text{if } \dot{\theta}_j(t) > 0 \\
b_j^c, & \text{if } \dot{\theta}_j(t) < 0
\end{cases}\\
\tau_j(t) = k_{\text{motor}} \left( k_p (\theta_j(t) - a_t) - k_d \dot{\theta_j(t)} \right)
\end{aligned}
\end{equation}
Here, $I_j$, $B_j$, and $b_j^c$ represent the joint inertia, viscous friction, and Static friction coefficients.
$\ddot{\theta}_j(t)$, $\dot{\theta}_j(t)$, $\theta_j(t)$, $\tau_j(t)$, and $f_j(t)$ denote the joint acceleration, joint velocity, joint position, joint torque from the PD controller, and Static friction at time $t$.
$k_{\text{motor}}$ is the motor strength factor, and $k_p$ and $k_d$ are the proportional and derivative gains of the PD controller.
Although static friction is typically greater than dynamic friction in reality, we assume they are equal for simplicity.

In conventional domain randomization, parameters such as $I_j$, $B_j$, $k_{\text{motor}}$, $k_p$, and $k_d$ are randomized.
From the dynamics equation, it is evident that $k_d$ and $B_j$ can cancel each other out, so only one of them needs to be randomized.
Additionally, Static friction $f_j(t)$ is typically not randomized in conventional approaches.
However, our experiments revealed that Static friction $f_j(t)$ significantly impacts the robot's performance.
To analyze the magnitude of $f_j(t)$, we performed parameter identification for the motor's physical parameters using the following method:
\begin{equation}
\label{equ:indentification}
\begin{gathered}
\theta_j^*(t) = A \sin(\omega t)\\
\min f(I_j, B_j, b_j^c) = \frac{1}{N} \sum_{i=0}^N \left( \theta_j^*(t_i) - \theta_j(t_i) \right)^2\\
\text{subject to } I_j > 0, \, B_j > 0, \, b_j^c > 0\\
\end{gathered}
\end{equation}

Here, $\theta_j^*(t)$ is the excitation signal, $A$ is its amplitude, $\omega$ is its frequency, $\theta_j(t_i)$ is the sampled joint angle, and $\theta_j^*(t_i)$ is the true value of the excitation signal.
We used the least squares method to identify $I_j$, $B_j$, and $b_j^c$.
The results of this parameter identification are presented in the Experiments section.

After parameter identification, we introduced Static friction $f_j(t)$ into the domain randomization process.
However, the inclusion of $f_j(t)$ introduced nonlinearity, significantly increasing training difficulty.
We observed that directly incorporating the identified Static friction $f_j(t)$ into the robot or manually compensating for it in the RL-generated actions $a_t$ caused the robot to remain stationary during training.
We hypothesize that this is due to the chaotic nature of early RL actions, with Static friction acting as a strong filter.
Fine-tuning a pre-trained model (trained without Static friction) by introducing $f_j(t)$ also failed, likely because the model weights were already trapped in local optima or saddle points, preventing effective gradient descent.

To train a robot capable of handling randomized Static friction $f_j(t)$, we developed two methods: the iterative method and the deception method.
In the iterative method, we fine-tuned the teacher model (trained without Static friction) multiple times, gradually introducing small amounts of $f_j(t)$.
However, models trained with this method exhibited severe jitter during real-world deployment, even making stable standing difficult.
As an alternative, we proposed the deception method, where we no longer aimed for perfect alignment between simulated and real-world Static friction.
Instead, we significantly expanded the randomization range of $f_j(t)$.
This approach successfully enabled real-world deployment.
We attribute the effectiveness of increased Static friction randomization to two factors:
1. During simulation training, increased randomization allows the robot to encounter scenarios with negligible Static friction, reducing training difficulty.
2. During Sim2Real, real-world joint Static friction may vary due to wear.
Expanding the randomization range enhances the robot's robustness to mechanical wear.

The parameters used in our domain randomization are listed in Table 2.
"Push velocity" refers to randomizing the base's linear velocity to simulate external pushes, eliminating the need to determine feasible push force magnitudes.

\begin{table}[]
\vspace{0.2cm}
\caption{DOMAIN RANDOMIZATION PARAMETERS}
\label{tab:randomization_params}
\centering
\begin{tabular}{@{}ccc@{}}
\toprule
\textbf{Parameter} & \textbf{Range} & \textbf{Unit} \\ \midrule
Joint armature & [0.8, 1.2] & multiplier \\
Joint damping & [0.8, 1.2] & multiplier \\
Joint Static friction & [0.0, 1.2] & multiplier \\
Kp & [0.95, 1.05] & multiplier \\
Motor strength & [0.8, 1.2] & multiplier \\
Ground friction & [0.2, 2.0] & multiplier \\
Payload & [-2, 3] & kg \\
Center of mass & [-0.25, 0.25] & m \\
Push interval & 8 & s \\
Push velocity & 1 & m/s \\
\bottomrule
\end{tabular}
\end{table}
\begin{figure}[h]
    \centering
    \includegraphics[scale=0.48]{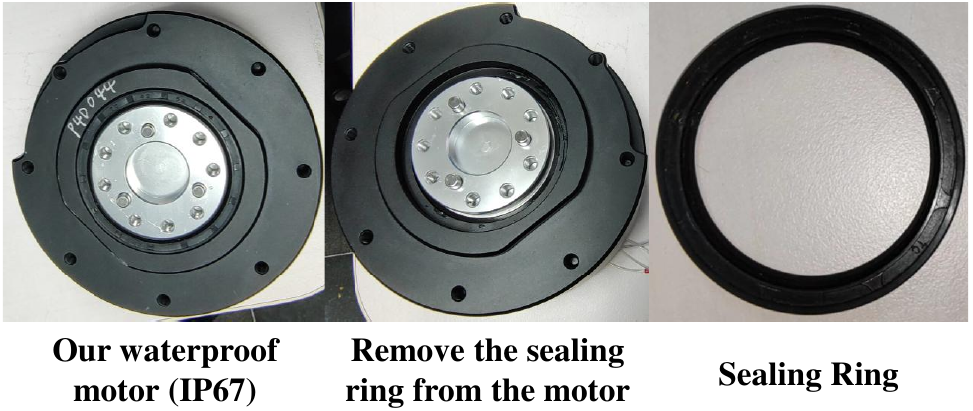}
    \caption{The motor used in Saturn Lite.}
    \label{fig:waterproof_motor}
\end{figure}
\begin{figure*}[h]
\centering
\vspace{0.5cm}
\includegraphics[scale=0.57]{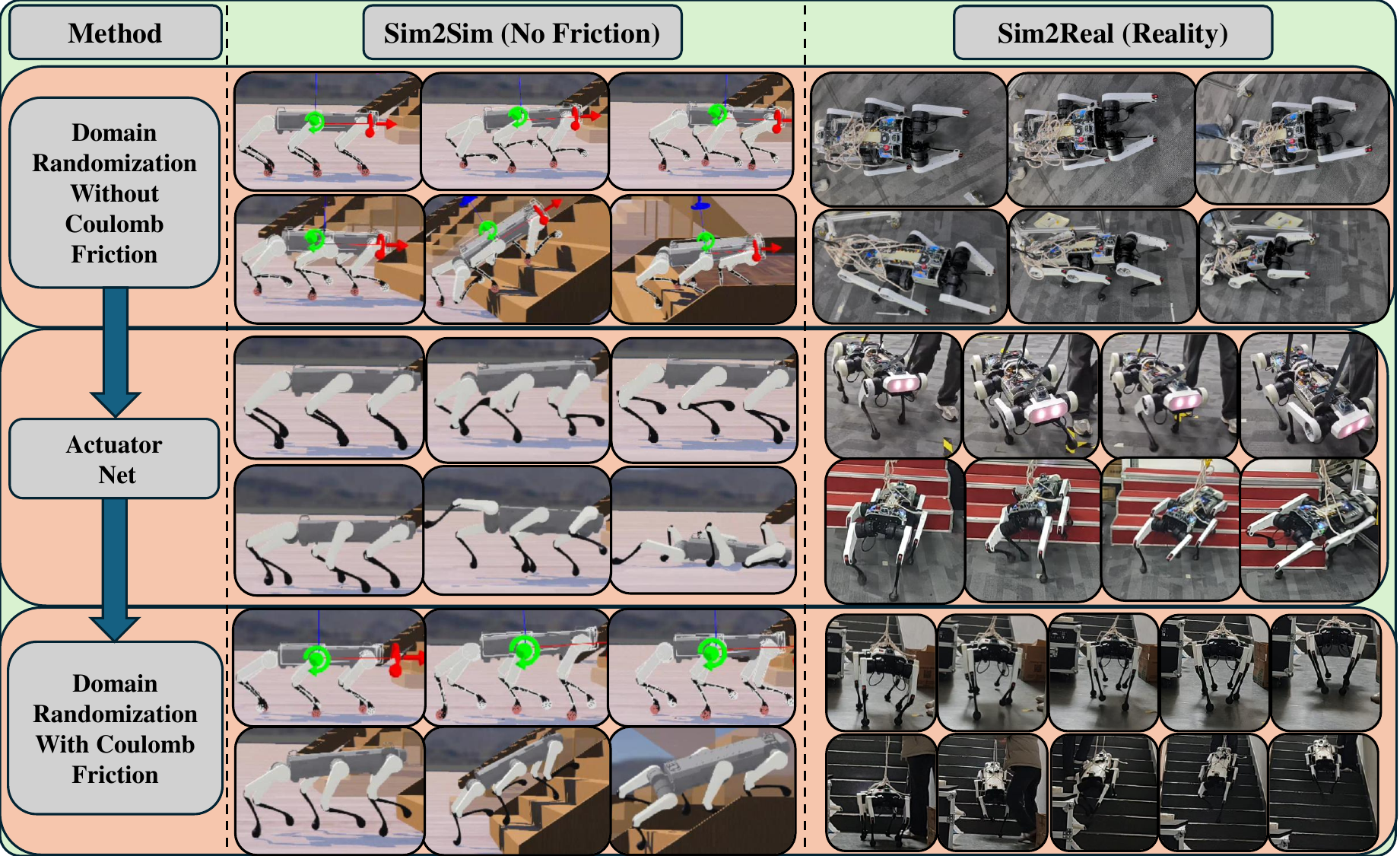}
\caption{Comparison of Sim2Sim and Sim2Real experiments.
The experiments focused on flat-ground walking and stair-climbing scenarios.}
\label{fig:sim2sim_sim2real_comparison}
\end{figure*}
\section{EXPERIMENTAL RESULTS}
In this section, we present the identification results of the physical parameters of the joints, explain why Static friction has a significant impact on our robot, and analyze which robots might face similar issues.
Additionally, we compare the performance of domain randomization without Static friction, Actuator Net, and domain randomization with Static friction in both Sim2Sim and Sim2Real scenarios.

\subsection{Measurement and Evaluation of Joint Equivalent Rotor Inertia, Damping, and Static Friction}
To identify the rotor inertia, damping, and Static friction of the joints, we used Equation \ref{equ:indentification} to estimate the physical parameters of the motors.
For simplicity, we focused on the shank motors of the Go1 and Saturn Lite robots under no-load conditions.
The identification results are summarized in Table \ref{tab:joint_parameters}.

    
    
    

\begin{table}[]
    \caption{SHANK MOTORS COMPARISON}
    \label{tab:joint_parameters}
    \centering
    \begin{tabular}{@{}llcc@{}}
    \toprule
    Property               & Type & Go1 & Saturn Lite (our) \\ 
    \midrule
    Inertia                & Mean   & $0.0121$ & $0.0145$ \\
    ($kg \cdot m^2$)       & Std    & $0.00223$ & $4.21\times10^{-4}$ \\
    \addlinespace[0.2cm]
    
    Viscous Friction       & Mean   & $0.0342$ & $0.0704$ \\
    ($N \cdot m \cdot s/rad$) & Std    & $0.00229$ & $0.0272$ \\
    \addlinespace[0.2cm]
    
    Static Friction       & Mean   & $0.0481$ & $0.442$\\
    ($N$)                  & Std    & $0.00299$ & $0.0661$ \\
    \addlinespace[0.2cm]
    
    \( f / \tau_{max} \) Ratio & Mean   & $0.13 \%$ & $0.98 \%$ \\
    (\%)                    & Std    & $5.70\times10^{-5}$  & $2.18\times10^{-4}$  \\
    \bottomrule
\end{tabular}
\end{table}

In most robotic reinforcement learning deployments, the impact of Static friction appears negligible.
We attribute this to the fact that, in most robots, Static friction constitutes a very small fraction of the joint torque, making it insignificant.
However, in our robot, Static friction accounts for a disproportionately large fraction of the joint torque, leading to a noticeable impact.
This is likely due to the waterproofing treatment and heavier link used in our robot, despite having similar torque capabilities.
The waterproof joint motor used in our robotic system, as shown in Figure \ref{fig:waterproof_motor}, exhibited a 70\% reduction in static friction through experimental removal of sealing rings.
To preserve the robot's functionality in applications, we retained the waterproof joint configuration with sealing rings during robot assembly, inherently maintaining the associated static friction effects.
To evaluate the effect of Static friction on robot performance, we conducted a series of comparative experiments in both Sim2Sim and Sim2Real scenarios.
The results are presented below.
\subsection{Comparison of Sim2Sim and Sim2Real with and without Static Friction}
We conducted comparative experiments using three different methods: Actuator Net, domain randomization (without Static friction), and domain randomization (with Static friction).
For the Sim2Sim experiments, we used Webots as the simulation platform.
The results of these experiments are illustrated in Figure \ref{fig:sim2sim_sim2real_comparison}.

In the Sim2Sim experiments, the robot achieved the best performance in Webots when using domain randomization without Static friction.
However, when using Actuator Net, the robot exhibited significant abnormal behavior, eventually leading to system failure.
With domain randomization that included Static friction, the robot displayed irregular jittering and jumping but was still capable of ascending and descending stairs.
This is primarily due to the RMA algorithm's ability to adapt to varying Static friction levels.

In the Sim2Real experiments, domain randomization without Static friction enabled the robot to walk backward on flat ground, but forward walking caused it to fall.
Using Actuator Net allowed the robot to walk forward, albeit unsteadily, and it was unable to climb stairs.
In contrast, domain randomization with Static friction achieved stable walking on flat ground and successful stair navigation.

The success of domain randomization without Static friction in Sim2Sim experiments indicates that our reinforcement learning training algorithm does not suffer from input-output interface configuration errors during deployment.
However, its failure in Sim2Real experiments suggests a significant discrepancy between simulation and reality.
The fact that Actuator Net performed well in Sim2Real flat-ground walking but failed in Sim2Sim  further supports the hypothesis that this discrepancy arises from differences in joint dynamics between simulation and reality.
Based on our earlier control-theoretic analysis, we hypothesized that Static friction is the primary cause of this discrepancy.
To validate this hypothesis, we implemented domain randomization with Static friction.
The experimental results demonstrated that this approach significantly improved Sim2Real performance, albeit with a slight sacrifice in Sim2Sim performance.
Given its strong adaptability to varying Static friction, we believe this method not only compensates for joint friction in robotic reinforcement learning but also serves as an effective means to counteract joint wear and fatigue.
\section{Conclusion}
Our experiments demonstrate that Static friction significantly impacts the Sim2Real of robotic reinforcement learning.
To mitigate this effect, we propose integrating Static friction into the training process.
Although the introduction of Static friction will cause the model to fail to train, we propose a simple trick to solve the problem.
Additionally, we hypothesize that using joints with lower Static friction-to-torque ratios could further alleviate this issue.

Over time, robots inevitably experience mechanical fatigue and wear, leading to changes in Static friction.
While RL models trained with randomized Static friction theoretically exhibit robustness against such degradation, our experimental constraints prevented direct validation of this hypothesis.
We believe this topic warrants further investigation.  


\bibliographystyle{IEEEtran}
\bibliography{IEEEabrv,my_ref}
\end{document}